\documentclass[10pt,twocolumn,letterpaper]{article}

\usepackage[pagenumbers]{wacv} 

\definecolor{wacvblue}{rgb}{0.21,0.49,0.74}
\usepackage[pagebackref,breaklinks,colorlinks,allcolors=wacvblue]{hyperref}
\usepackage{multirow}
\usepackage{graphicx}
\usepackage{comment}
\usepackage{xcolor}  

\usepackage[accsupp]{axessibility}  

\def\wacvPaperID{7, PRAW} 
\def\confName{WACV}
\def\confYear{2026}

\title{From Pixels to Purchase: Building and Evaluating a Taxonomy-Decoupled Visual Search Engine for Home Goods E-commerce}

\author{
Cheng Lyu \quad Jingyue Zhang \quad Ryan Maunu \quad Mengwei Li \quad Vinny DeGenova \quad Yuanli Pei\\
Wayfair\\
{\tt\small \{clyu1, jzhang2, rmaunu, mli2, vdegenova, ypei\}@wayfair.com}
}

\begin{document}
\maketitle
\begin{abstract}
Visual search is critical for e-commerce, especially in style-driven domains where user intent is subjective and open-ended. Existing industrial systems typically couple object detection with taxonomy-based classification and rely on catalog data for evaluation, which is prone to noise that limits robustness and scalability. We propose a taxonomy-decoupled architecture that uses classification-free region proposals and unified embeddings for similarity retrieval, enabling a more flexible and generalizable visual search. To overcome the evaluation bottleneck, we propose an LLM-as-a-Judge framework that assesses nuanced visual similarity and category relevance for query-result pairs in a zero-shot manner, removing dependence on human annotations or noise-prone catalog data. Deployed at scale on a global home goods platform, our system improves retrieval quality and yields a measurable uplift in customer engagement, while our offline evaluation metrics strongly correlate with real-world outcomes.
\end{abstract}    
\section{Introduction}

Visual search has emerged as a critical capability in e-commerce, particularly in aesthetic and style-driven domains such as home goods and fashion.
Unlike text-based queries, which require users to articulate their intent precisely, visual search enables discovery through images, videos, or multimodal inputs.
This capability is essential to capture user intent.
For example, a user who queries with a photo of a distinctive chair may seek not only an exact replica, but also related items that capture its ``mid-century modern'' or ``rustic'' qualities.
Developing systems that can effectively interpret and respond to such subjective and open-ended intent is both a research challenge and a commercial imperative.

The dominant paradigm for commercial visual search, pioneered by leading e-commerce and content discovery platforms, follows a multi-stage pipeline: an object detector localizes prominent products in a query image, which are then encoded into a feature space for similarity search \cite{jing2015visual, yang2017visual}.
Subsequent industry work has refined this approach through more sophisticated ranking methods and web-scale optimizations \cite{hu2018web, zhang2018visual}.
This general workflow has shown some effectiveness, but its conventional implementation has several limitations.

Standard visual search systems, including our prior-generation implementations, have historically relied on object detectors that both localize and classify items under a domain-specific taxonomy \cite{yamaguchi2013paper, bell2015learning, jing2015visual, yang2017visual, hu2018web, zhang2018visual}.
For instance, a home goods company might define product category taxonomies to differentiate \textit{Sofas} and \textit{Coffee Tables}.
While such distinctions are straightforward, taxonomies often introduce ambiguous category boundaries and miss important visual nuances.
As a result, visually similar products, such as \textit{Sofas} and \textit{Loveseats}, may be forced into separate product categories despite substantial overlap in appearance.
This fine-grained categorization often drives up training data costs and impedes adaptability to taxonomy updates, ultimately leaving systems fragile and expensive to scale in practice.

In addition, a reliable framework for measuring system performance is lacking.
Prior work often relies on offline evaluation, comparing outputs to a presumed ``ground truth'' using metrics such as classification accuracy or retrieval relevance \cite{jing2015visual, yang2017visual, hu2018web, zhang2018visual}.
Such methods often favor exact matches and overlook a user's real intent, \ie, finding visually similar results, and are fundamentally misaligned with the goals of visual search.
This issue is compounded by noisy and incomplete catalog data, which means the evaluation reflects flawed labels rather than true visual quality.
Human annotations have been the fallback to address such issues.
The process is typically slow, costly, and still suffers from the same noisy metadata issue.

To address these limitations, we introduce the following contributions:
\begin{enumerate}

\item \textbf{Taxonomy-decoupled architecture.}
We present a visual search system that decouples object detection from fine-grained classification.
Instead of relying on large multimodal models and text prompts, the system uses an object detector trained on groups of visually similar categories only to generate region proposals.
This design eliminates the dependency on rigid taxonomies while preserving the low latency performance required for deployment at industrial scale.
Detected objects are then encoded into a shared embedding space, where approximate nearest-neighbor search retrieves visually similar items without explicit classification.
This taxonomy-decoupled architecture enables open-ended visual search driven by appearance similarity to the query image.

\item \textbf{Scalable LLM-as-a-Judge Evaluation Framework.}
We develop a novel Large Language Model (LLM)-as-a-Judge evaluation framework that assesses true visual similarity at scale without relying on a dataset of human annotations.
It leverages a state-of-the-art LLM to serve as a reliable proxy for human judgment by evaluating the category relevance and visual similarity between a query and the retrieved results.

\end{enumerate}

We deployed our proposed system at Wayfair, demonstrating improved retrieval quality and measurable uplifts in live customer metrics.
Furthermore, our evaluation metrics demonstrate a strong correlation with customer outcomes, providing a statistically validated link between offline evaluation and real-world industrial impact.
\section{Related Work}

Industrial visual search has progressed in three directions: large-scale commercial visual search systems leveraging object detection and similarity retrieval, unified embedding frameworks for visual similarity retrieval, and recent advances using LLMs for zero-shot search and evaluation tasks.
In the following sections, we discuss representative work in each area.

\subsection{Commercial Visual Search Systems}

Commercial visual search systems, particularly in e-commerce and content discovery platforms \cite{jing2015visual, yang2017visual}, are commonly built around a multi-stage process that detects objects within an image, encodes them into vector representations, and retrieves similar items from the catalog via nearest-neighbor search.
Subsequent industry efforts have expanded this paradigm by scaling to increasingly large catalogs and incorporating advances such as deep learning, attention mechanisms, and multimodal ranking \cite{zhang2018visual, hu2018web, du2022amazon}.
For instance, query-dependent class prediction has been shown to improve retrieval relevance \cite{hu2018web}, while fusion methods that combine model and search-based approaches have been applied for joint detection and feature learning \cite{zhang2018visual}.
Other systems, such as \textit{Shop the Look}, employ a combination of product localizer, fine-grained classifier, and feature extractor to bridge the domain gap between query images and product images in the catalog \cite{du2022amazon}.
Despite their effectiveness at scale, these approaches generally preserve a tight coupling between object detection and classification, constrained by predefined catalog taxonomies.

\subsection{Unified Embedding Frameworks}

A critical step in modern visual search is learning a unified embedding space that can represent diverse items.
Early work \cite{zhai2019learning} demonstrated this through a multi-task framework that trained a single model across multiple product inventories, enabling cross-category visual similarity search and reducing the need for separate category-specific models.
Subsequent approaches, such as PinnerSage \cite{pal2020pinnersage} and OmniSearchSage \cite{agarwal2024omnisearchsage}, extended this idea to multimodal embeddings by incorporating user behavior and textual data, thereby supporting broader retrieval tasks across users, queries, and content.
Despite these advances, such frameworks generally remain grounded in visual representations that depend on underlying category structures for initial item identification and organization.
While multimodal foundation models like CLIP \cite{radford2021learning} offer a potential solution by leveraging natural language pre-training, they typically require additional fine-tuning to support enterprise visual search systems \cite{zhu2024bringing}.

\subsection{LLMs in Search and Evaluation}

The application of LLMs is rapidly transforming multiple aspects of search, including query understanding, result generation, and relevance assessment \cite{xiong2024search}.
In large-scale e-commerce platforms, recent systems \cite{wang2024improving, shang2025knowledge} have shown that LLMs can generate nuanced relevance judgments for complex query-product pairs.
These judgments are then used to improve re-ranking algorithms or to provide high-quality supervision for training smaller, more efficient production models.
While these approaches utilize LLMs within search or ranking pipelines to refine relevance signals, they generally do not tackle the zero-shot, scalable evaluation of core visual search system's output, such as subjective visual similarity or upstream component quality, in a judge-like capacity.
\section{Visual Search Architecture}

\begin{figure*}[htbp]
 \centering
 \includegraphics[width=0.96\textwidth]{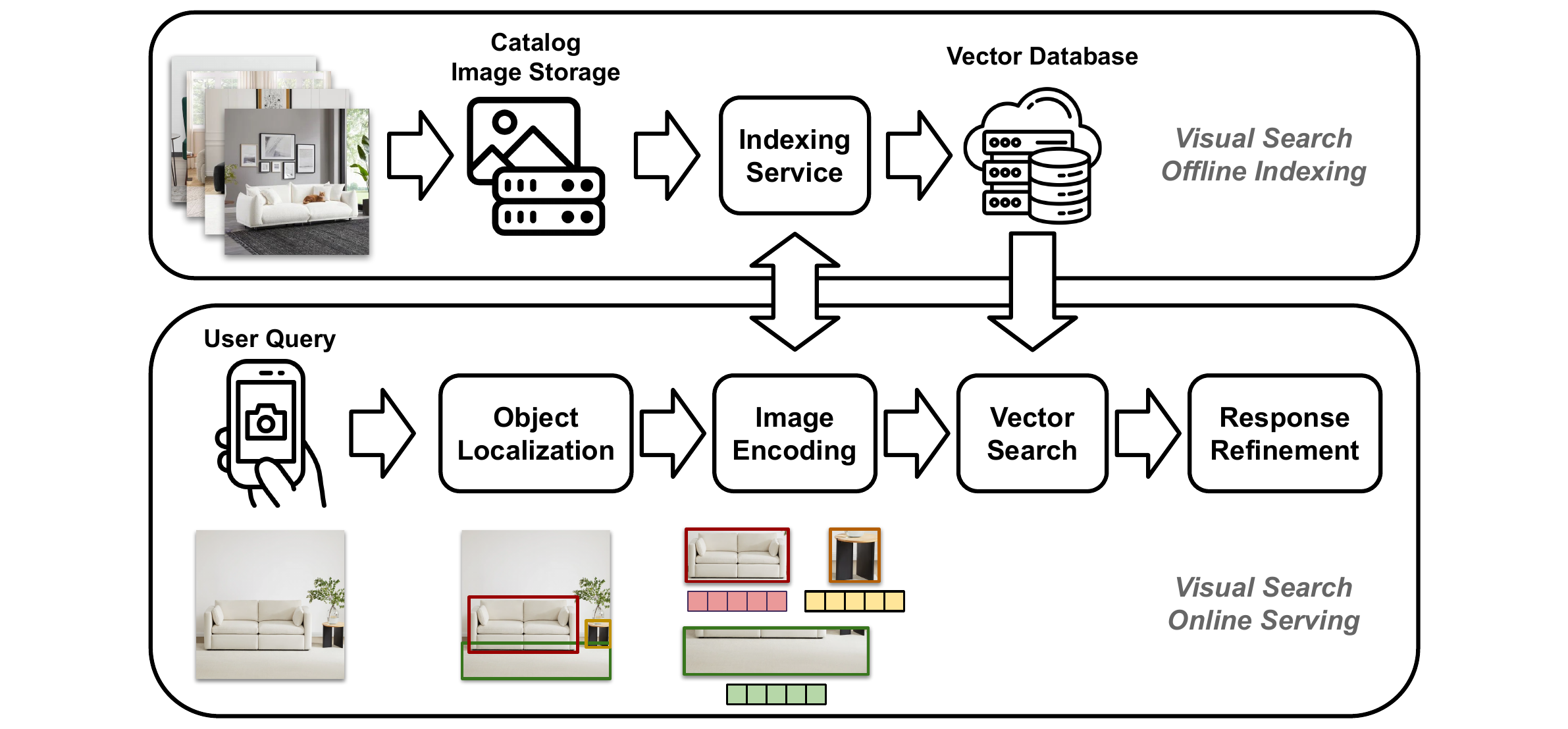}
 \caption{An overview of our taxonomy-decoupled visual search architecture. 1) \textit{Offline Indexing:} Product images from the catalog are encoded into unified embeddings to build a large-scale vector index. 2) \textit{Online Serving:} A taxonomy-decoupled object detector first localizes objects in the user's query image. Each detected object is then encoded and used to query the vector index. The retrieved candidates are subsequently processed through a multi-stage refinement pipeline to generate a shoppable gallery.}
 \label{fig:architecture}
\end{figure*}

We introduce a visual search architecture that decouples object localization from classification.
As shown in \Cref{fig:architecture}, the system is composed of two main processes: \textbf{Offline Indexing}, where a unified vector index of product images is built, and \textbf{Online Serving}, where live user queries are processed.
The online pipeline begins with a taxonomy-decoupled object detector identifying items in a user's query image.
Each detected item is then encoded into an embedding vector to search the index for visually similar products.
Finally, a post-retrieval re-ranking stage organizes the results into a shoppable gallery for the user.
The following sections detail each component of this architecture.

\subsection{Taxonomy-Decoupled Object Localization}
\label{sec:od}

One key component in our online pipeline is an object detector that localizes all potentially sellable products within a user's query image, ranging from clean product shots to cluttered, multi-object scenes.
The detector is trained offline and deployed online to support real-time inference on user queries.

\textbf{Offline Development.}
To decouple localization from a rigid catalog taxonomy, we developed an object detector based on the one-stage, anchor-free YOLOX architecture \cite{ge2021yolox}.
We trained the detector on the basis of \textit{superclasses} which group thousands of fine-grained product categories into a few hundred visually similar groups.
For example, visually similar product categories such as \textit{Sofas} and \textit{Loveseats} are merged into a single superclass.
By simplifying categories and annotations, this approach enables efficient collection of training data (at the superclass level) for a robust localizer that handles diverse visual patterns, without requiring fine-grained taxonomies at inference.

\textbf{Online Serving.}
In the production pipeline, YOLOX generates object proposals from the user's query image.
Low-confidence detections are removed, and (super)class-agnostic Non-Maximum Suppression (NMS) merges redundant boxes.
By discarding predicted superclass labels, the system can localize out-of-distribution (OOD) objects and defer semantic understanding to the retrieval stage, which matches products primarily by visual appearance.
For images containing multiple objects, detections are ranked by a weighted combination of confidence and relative bounding-box area, serving as a proxy for user intent and determining retrieval order.

\subsection{Unified Embedding for Similarity Search}
\label{sec:cammo}

Our system relies on a unified embedding space to represent both catalog images and query objects for similarity search in offline indexing and online serving.
This representation captures both high-level categorical features and fine-grained visual details without relying on explicit object superclass prediction.

\textbf{Offline Development.}
We developed our embedding model by fine-tuning an open-source variant of the OpenCLIP architecture \cite{radford2021learning}.
Specifically, we started with the OpenCLIP-H/14 model \cite{cherti2023reproducible} pre-trained on the Datacomp dataset \cite{gadre2023datacomp}.
We fine-tuned this model using tens of millions of product images across hundreds of product categories.
This yielded a robust visual representation that supports both broad category understanding and fine-grained instance retrieval, forming the foundation of our taxonomy-decoupled visual search system.

\textbf{Offline and Online Usage.}
The resulting embedding model is used in both the offline index building and online serving processes.
Offline, we first use the model to encode over 200 million images in our product catalog, generating the embeddings that populate the vector index.
Online, the same model is deployed as a service to encode each object patch identified by the detector in a user's query image, generating the query vectors for retrieval.

\subsection{Offline Indexing and Online Retrieval}
\label{sec:vvs}
Another key system component is a large-scale indexing and retrieval system designed for real-time performance. 

\textbf{System Foundation.}
We leverage Google's Vertex AI Vector Search, a managed service built upon the Scalable Nearest Neighbors (ScaNN) algorithm \cite{guo2020accelerating}.
This service provides a unified global search space that effectively supports (super)class-agnostic queries across our entire catalog.
By leveraging streaming ingestion, we incorporate the newly added products into the index within hours, ensuring index freshness.
Additionally, ScaNN's anisotropic vector quantization offers an effective balance between search speed, memory footprint, and recall.

\textbf{Offline and Online Usage.}
In the offline indexing stage, we populate the vector search index with embeddings of our catalog images.
In addition to image embeddings, we tag each vector with metadata from the catalog, such as geographic identifiers corresponding to country-specific inventory.
This metadata has been found to be beneficial both in refining the retrieval process and in improving the results returned by vector search.

During the online stage, our retrieval pipeline adopts a multi-stage filtering strategy to balance efficiency and retrieval quality.
At the index level, queries first leverage metadata tags (\eg, geographic availability) to constrain the search space, thereby reducing unnecessary candidate exploration and improving latency.
Following vector search, a second application-layer filtering stage applies domain-specific constraints, such as ensuring product findability and catalog compliance. 
Such filtering mechanisms, supported by rich catalog metadata, enable the system to efficiently prune irrelevant candidates, reduce computational overhead, and yield contextually relevant results.

\subsection{Response Refinement}
\label{sec:proxy}

The final stage of our online pipeline transforms the raw vector search results into a customer-facing product gallery.
This stage employs a multi-step re-ranking process designed to optimize for visual relevance, diversity, and shoppability.

First, the pipeline refines raw results by resolving redundancies.
This step is especially critical given the prevalence of duplicate products in large catalogs.
Since an effective vector search system may readily surface such duplicates, we apply deduplication at both the image and product levels to preserve diversity without compromising relevance.

Next, the pipeline dynamically assigns a \textit{broad class} label to the results for each detected object.
Unlike the visual superclasses in \Cref{sec:od}, this label is derived from business logic to ensure shoppability.
For example, visually distinct yet functionally related product categories like \textit{Indoor Chaise Lounges} and \textit{Loveseats} are aggregated under the customer-facing broad class ``Sofas''.
Inferred from the metadata of top-retrieved candidates, this step effectively \textit{defers} taxonomy constraints rather than eliminating them, allowing the search to remain flexible and data-driven while still aligning with merchandising needs.

Finally, results are filtered to include only shoppable items meeting business requirements.
For multi-object queries, a round-robin strategy ensures balanced representation across detected objects, preventing any single object from dominating the gallery.
Overall, this multi-step filtering and re-ranking process not only refines the raw vector search output but also provides the flexibility to incorporate additional domain-specific ranking signals, ultimately yielding a practical and engaging shoppable experience.
\begin{figure*}[htbp]
  \centering
  \includegraphics[width=\textwidth]{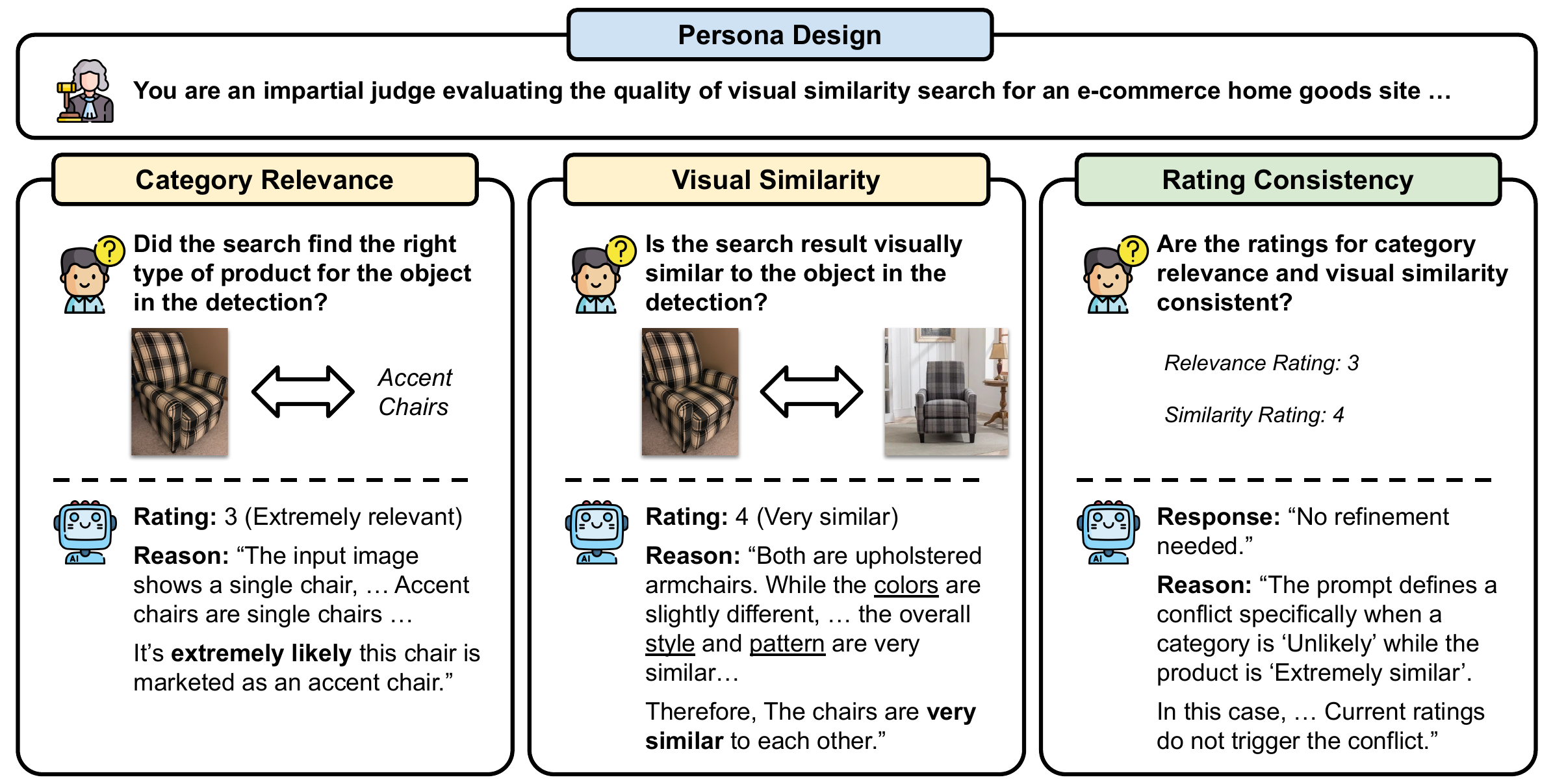}
  \caption{LLM-as-a-Judge evaluation framework evaluates each query-result pair in three steps: 1) \textit{Category Relevance} rating whether the retrieved item belongs to the same product category as the query on a 3-point scale; 2) \textit{Visual Similarity} rating the aesthetic and functional similarity between query and result on a 5-point scale; and 3) \textit{Rating Consistency} verifying logical consistency between the two ratings and refining judgments when predefined conflicts arise (\eg, low category relevance but high visual similarity).}
  \label{fig:llmaj}
\end{figure*}

\section{LLM-as-a-Judge: A Validated Proxy for Human Evaluation}
\label{sec:llm_judge_framework}

A key bottleneck in improving style-driven visual search systems is the lack of a scalable evaluation framework.
To overcome this bottleneck, we introduce and validate a novel evaluation framework using LLM-as-a-Judge, which serves as a scalable proxy for human evaluation.
This section details the framework's design and validates its reliability, establishing it as the basis for our subsequent experiments.

\subsection{Framework Definition}
As illustrated in \Cref{fig:llmaj}, our framework assigns an LLM the role of an impartial judge within the application domain, \eg, an e-commerce home goods platform.
Using a multi-task prompting strategy, the model evaluates each query-result pair on two tasks, \textit{Category Relevance} and \textit{Visual Similarity}, and produces both numerical ratings and textual justifications for each assessment.
The framework further incorporates a \textit{Rating Consistency} step that reconciles and refines judgments when conflicts arise.

The \textit{Category Relevance} task prompts the model to evaluate functional compatibility on a 3-point scale, guiding it to accept closely related categories while penalizing distinct functional mismatches (\eg, a straight-line \textit{Sofa} \vs an L-shaped \textit{Sectional}).
This evaluation addresses the fundamental user intent rooted in product categories.
Although our search architecture is taxonomy-decoupled to broaden the retrieval scope, this validation is crucial for ensuring that search flexibility does not come at the cost of the product's intended purpose or the user's functional expectations.

The \textit{Visual Similarity} task focuses on capturing the aesthetic and stylistic nuances that define the core objective of visual search.
It prompts the model to quantify the likeness between the query and result based on fine-grained attributes such as color and material, using a 5-point Likert scale ranging from \textit{Extremely similar} to \textit{Extremely different}.
The evaluation is framed to assess how faithfully the result captures the specific design and appearance of the object detected in the input image.
This granular metric is essential for differentiating between high-quality matches and merely passable results, providing a more accurate reflection of user satisfaction than binary classification.

The last \textit{Rating Consistency} step ensures logical coherence between the Category Relevance and Visual Similarity ratings.  
This step prompts the model to self-correct in cases of conflicting ratings, such as when a result is judged extremely similar but belongs to an irrelevant category. 
Such scenarios are logically inconsistent, and the model is instructed to revisit its ratings to resolve the discrepancy. 
This refinement occurs within the model's internal reasoning to ensure the reliability of its judgments, and it is omitted from the final output.

\subsection{LLM-as-a-Judge vs Human Evaluation}

\begin{table*}[htbp]
\centering
\resizebox{0.9\textwidth}{!}{
\begin{tabular}{l c c c c c}
\toprule
& \multicolumn{3}{c}{\textbf{Ordinal Reliability Metrics}} & \multicolumn{2}{c}{\textbf{Binary Decision Metrics}} \\
\cmidrule(lr){2-4} \cmidrule(lr){5-6}
\textbf{Evaluation Task} & Weighted Kappa ($\kappa_w$) & Spearman's $\rho$ & MAE & F1-Score & MCC \\
\midrule
Category Relevance
& $0.811 \pm 0.036$
& $0.841 \pm 0.031$
& $0.173 \pm 0.033$
& $0.928 \pm 0.010$
& $0.781 \pm 0.036$ \\
Visual Similarity
& $0.894 \pm 0.026$
& $0.945 \pm 0.015$
& $0.251 \pm 0.051$
& $0.955 \pm 0.010$
& $0.878 \pm 0.028$ \\
\bottomrule
\end{tabular}%
}
\caption{LLM-as-a-Judge Validation. Inter-rater reliability analysis against human ratings (N=255, values as mean $\pm$ standard error) shows strong agreement on both ordinal and binarized metrics, validating its use as a scalable and nuanced proxy for human judgment.}
\label{tab:llm_judge_metrics}
\end{table*}

We evaluated the reliability of the LLM-as-a-Judge against human expert ratings using a curated set of 1k query-result pairs, designed to capture challenging cases where the exact item might not exist in the catalog. 
For a primary-object subset (N=255) reported in \Cref{tab:llm_judge_metrics}, the task emphasizes nuanced similarity judgments rather than exact matches.

Agreement between the LLM-as-a-Judge and human expert ratings was measured using two groups of metrics.
To assess the model's ability to capture the fine-grained degrees of likeness in a style-driven domain, we used ordinal reliability metrics on the original scales (3-point Category Relevance, 5-point Visual Similarity): Mean Absolute Error (MAE), Quadratic Weighted Kappa ($\kappa_w$) \cite{cohen1968weighted}, and Spearman's Rank Correlation ($\rho$).
To reflect practical business cases where a customer's decision is often binary (\ie, similar enough or not), we also included binary decision metrics by binarizing the 3-point or 5-point ratings into \textit{Is Relevant} and \textit{Is Similar} and evaluating with F1-Score and Matthews Correlation Coefficient (MCC) \cite{matthews1975comparison}.

As shown in \Cref{tab:llm_judge_metrics}, LLM-as-a-Judge achieves strong agreement with human experts on primary objects.
On Visual Similarity, it attains high $\kappa_w$ 0.894 (\textit{almost perfect agreement}), high Spearman's $\rho$, low MAE.
The agreement on Category Relevance is slightly lower than Visual Similarity but still shows \textit{substantial agreement} with $\kappa_w$ 0.811. 
High F1 and MCC scores from the binary decision metrics for both tasks further demonstrate the model's effectiveness in capturing human intent regarding retrieval relevance and similarity. Overall, these results confirm the framework's reliability and establish it as a well-calibrated surrogate for human judgment.
\section{Visual Search System Evaluation}
\label{sec:experiments}

Leveraging the validated LLM-as-a-Judge framework (\Cref{sec:llm_judge_framework}), we evaluate our visual search system by benchmarking its overall performance against relevant baselines and analyzing the performance of each key component.

\subsection{End-to-End System Performance}
\label{sec:e2e_results}

Our primary benchmark used 1k images sampled from multiple sources: user query images, product images, and AI-generated inspirational images used in the live platform.
These images were held out from the training sets of both the detector and embedding models to prevent data leakage.
We benchmarked the following visual search systems:
\begin{itemize}
    \item \textbf{Taxonomy-decoupled}: The proposed visual search system, which uses (super)class-agnostic object detection and a unified embedding for retrieval.
    \item \textbf{Class-dependent}: Our legacy taxonomy-dependent system, which relies on a fine-tuned Inception-ResNet \cite{szegedy2017inception} embedding model where retrieval is guided by an explicit classification stage.
    \item \textbf{Google Lens}: We tasked human agents with using the Google Lens application to manually input the detected objects in the images into Google Lens along with the e-commerce company name in the text prompt.
\end{itemize}
We used four key metrics for evaluation: \textit{Relevance Precision} (Rel P@k), the fraction of top-$k$ results from the correct product category;
\textit{Visual Similarity Precision} (VS P@k), the fraction of visually similar results;
\textit{Success Rate} (Success@k), the percentage of queries with at least one relevant and visually similar result;
and \textit{nDCG@k}, to measure ranking quality.
All metrics were calculated based on the evaluation from the LLM-as-a-Judge framework.

As detailed in \Cref{tab:main_results}, the benchmark results for the 1k-image set confirm that our proposed system significantly outperforms both the commercial and legacy baselines.
The most pronounced advantage is in fine-grained matching, where our system's ability to capture nuanced aesthetic details leads to a substantial lift in Visual Similarity Precision. 
Our system achieves a much higher Success Rate, demonstrating that its taxonomy-decoupled architecture not only simplifies the system by removing the rigid classification dependency but also yields a significant improvement in retrieval quality.

Additionally, to assess scalability, we evaluated the proposed taxonomy-decoupled system on a 15k image set and observed that the performance remains high (Visual Similarity P@6 of 47.5\% and Success Rate@6 of 83.0\%)\footnote{Note that we did not evaluate Google Lens on the larger 15k image dataset, as the manual effort would be prohibitive.}.
This demonstrates the system's robust generalization across a broader distribution of real-world queries.

\begin{figure*}[htbp]
  \centering
  \includegraphics[width=\textwidth]{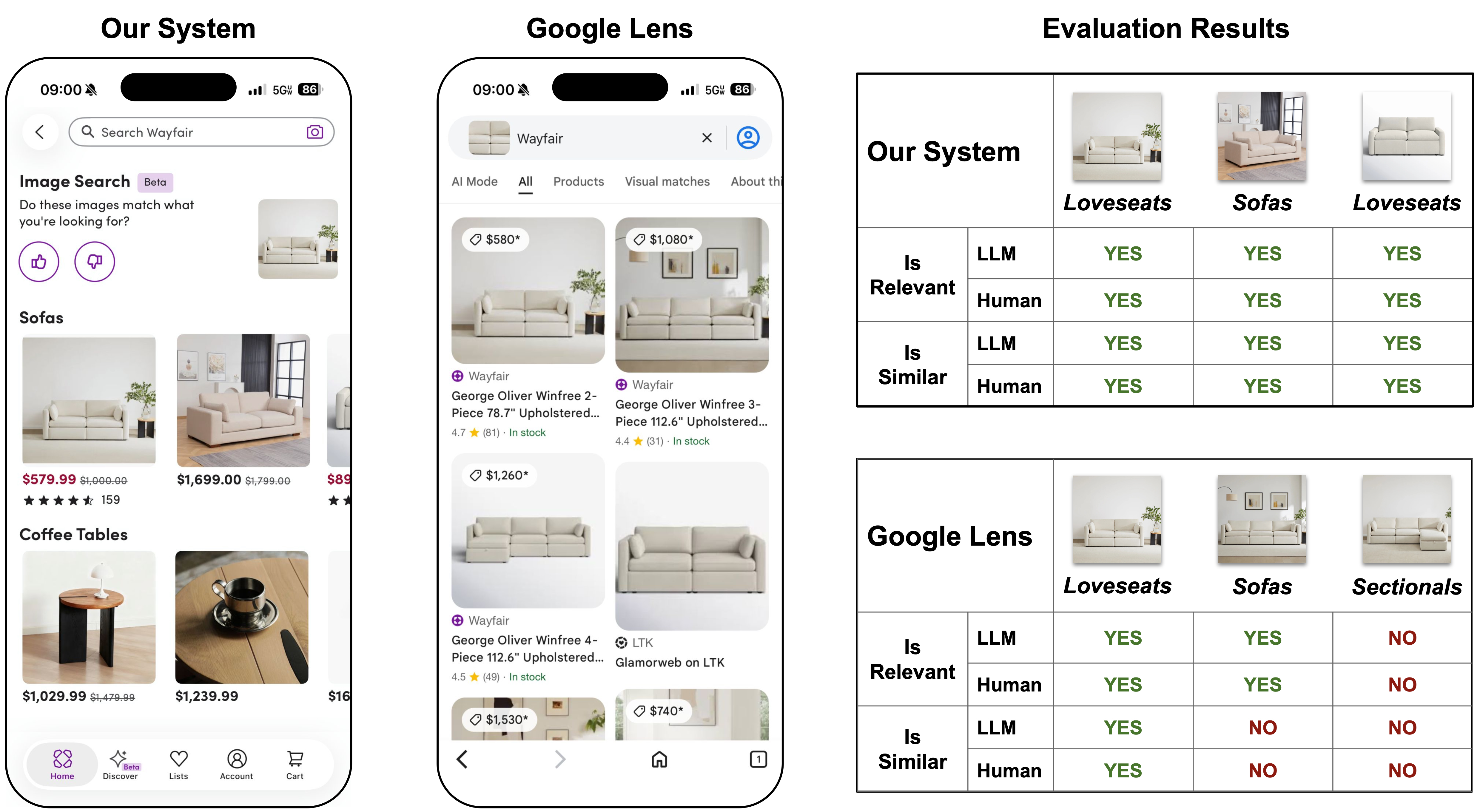}
  \caption{Qualitative comparison. Left: Our system detects multiple objects (``Sofas'', ``Coffee Tables''). For the primary object, it retrieves similar products from product categories \textit{Loveseats} and \textit{Sofas}, organized under a \textit{broad class} ``Sofas'' (\Cref{sec:proxy}). Center: Google Lens results. Right: Evaluation using granular product categories. Human validation of the LLM-as-a-Judge assessment confirms that retrieving \textit{Loveseats} for a sofa query is correct due to their shared straight-line configuration, while penalizing \textit{Sectionals} (from Google Lens) for their incompatible L-shaped structure.}
  \label{fig:qualitative_comparison}
\end{figure*}

\Cref{fig:qualitative_comparison} qualitatively demonstrates how our evaluation balances visual similarity with category-aligned shopping intent.
Given a query image of a cream-colored, straight-line upholstered sofa, our system retrieves visually consistent results from distinct granular product categories: \textit{Loveseats} and \textit{Sofas}.
This cross-category retrieval is evaluated as correct because both specific categories belong to the same broad class (``Sofas'') and share the query's critical straight-line configuration.
In contrast, Google Lens returns an item from product category \textit{Sectional}.
Although related, it is penalized in our evaluation because the L-shaped configuration fundamentally contradicts the query's visual structure and functional intent.
This distinction clarifies that our framework does not indiscriminately reward cross-category diversity, but rather prioritizes results that align with the user's specific visual requirements, regardless of their granular category label.

\begin{table}[tbp]
\centering
\resizebox{\columnwidth}{!}{%
\begin{tabular}{c l c c c c}
\toprule
\textbf{$k$} & \textbf{Method} & \textbf{Rel P@k} & \textbf{VS P@k} & \textbf{Success@k} & \textbf{nDCG@k} \\
\midrule
\multirow{3}{*}{1}
& Google Lens       & 86.4 & 41.2 & 41.1 & 1.000 \\
& Class-dependent   & 82.5 & 43.5 & 43.2 & 1.000 \\
& \textbf{Taxonomy-decoupled}    & \textbf{94.4} & \textbf{59.3} & \textbf{58.9} & 1.000 \\
\cmidrule(lr){1-6}
\multirow{3}{*}{3}
& Google Lens       & 84.5 & 37.4 & 59.4 & 0.947 \\
& Class-dependent   & 82.1 & 40.4 & 59.1 & \textbf{0.956} \\
& \textbf{Taxonomy-decoupled}    & \textbf{93.6} & \textbf{54.5} & \textbf{77.4} & 0.950 \\
\cmidrule(lr){1-6}
\multirow{3}{*}{6}
& Google Lens       & 76.7 & 31.4 & 66.2 & 0.928 \\
& Class-dependent   & 82.1 & 38.2 & 67.0 & \textbf{0.932} \\
& \textbf{Taxonomy-decoupled}    & \textbf{92.9} & \textbf{51.1} & \textbf{85.3} & 0.926 \\
\bottomrule
\end{tabular}%
}
\caption{End-to-end system benchmark on 1k image test set. Comparison of the taxonomy-decoupled system, Google Lens, and the legacy class-dependent system.}
\label{tab:main_results}
\end{table}

\subsection{Component-Level Evaluation}
\label{sec:component_analysis}

The strong end-to-end performance of the system is grounded in the quality of its foundational components. Next, we evaluate the effectiveness of the object detector and the embedding model individually.

\subsubsection{Class-Agnostic Object Localization Evaluation}
\label{sec:detection_analysis}

We benchmarked our in-house YOLOX-based detector against the Google Vision API on 29,777 catalog images shot in different room settings with multiple products.
To reflect distinct user scenarios, we designed two (super)class-agnostic evaluation settings:
1) \textit{All Objects}, evaluating all ground-truth objects to assess performance in cluttered scenes, and
2) \textit{Prominent Object}, comparing the single largest ground-truth object against the best prediction (ranked by size-weighted confidence) to simulate a primary product search. For each setting, we report three metrics: mean average precision (mAP), precision, and recall.

As shown in \Cref{tab:detection_results}, our custom detector substantially outperforms the commercial baseline.
On the Prominent Object task, our model achieves an mAP of 53.3\%, compared to 26.7\% for Google Vision API, demonstrating superior ability to pinpoint the user's primary item of interest.
Furthermore, its recall of 70.1\% on the All Objects task (\vs~30.3\%) shows its improved capability to identify all potentially sellable items, which is critical to customer discovery.

\begin{table}[htbp]
\centering
\resizebox{\columnwidth}{!}{
\begin{tabular}{l l c c}
\toprule
\textbf{Setting} & \textbf{Metric} & \textbf{Google Vision (\%)} & \textbf{Our Model (\%)} \\
\midrule
\multirow{3}{*}{\begin{tabular}{@{}l@{}}Prominent \\ Object\end{tabular}} 
  & mAP       & 26.7 & \textbf{53.3} \\
  & Precision & 51.1 & \textbf{71.5} \\
  & Recall    & 48.8 & \textbf{70.7} \\
\cmidrule(lr){1-4}
\multirow{3}{*}{All Objects} 
  & mAP       & 18.2 & \textbf{47.5} \\
  & Precision & 53.3 & \textbf{57.7} \\
  & Recall    & 30.3 & \textbf{70.1} \\
\bottomrule
\end{tabular}
}
\caption{Comparison between (super)class-agnostic object detection and Google Vision, evaluated on both all detected objects and prominent objects.}
\label{tab:detection_results}
\end{table}

\subsubsection{Embedding Model Quality}
\label{sec:embedding_analysis}

\begin{table}[tp]
\centering
\resizebox{\columnwidth}{!}{
\begin{tabular}{lcc}
\toprule
\textbf{Model} & \textbf{Class Acc.@1 (\%)} & \textbf{Recall@1 (\%)} \\
\midrule
Inception-ResNet (Our Legacy) & 52.0 & 42.1 \\ 
\begin{tabular}{@{}r@{}}CLIP ViT-B/16\\ 
\end{tabular} & 62.9 & 27.7 \\
\begin{tabular}{@{}r@{}}CLIP ViT-H/14\\ 
\end{tabular} & 77.5 & 41.8 \\
\begin{tabular}{@{}r@{}}\textbf{CLIP ViT-H/14 ft (Our New)}\\ 
\end{tabular} & \textbf{79.6} & \textbf{56.4} \\
\bottomrule
\end{tabular}
}
\caption{Comparison of embedding model performance on large-scale Category Classification and Exact Product Retrieval benchmarks.}
\label{tab:embedding_results}
\end{table}

To evaluate the embedding model performance, we benchmarked our fine-tuned OpenCLIP ViT-H/14 model against several baselines, including a fine-tuned Inception-ResNet model used in our legacy taxonomy-dependent system, and two common off-the-shelf CLIP model variants, \textit{CLIP ViT-H/14} and \textit{CLIP ViT-B/16}.

We perform two evaluation tasks using distinct datasets sampled from our catalog:
1) \textit{Category Classification}, using 50k images across 400 classes to measure linear probe Top-1 accuracy and assess broad semantic understanding, and
2) \textit{Exact Product Retrieval}, using 280k image pairs of different scenes sourced from the same product across 320 classes to measure Recall@1 and evaluate fine-grained instance identification.

As shown in \Cref{tab:embedding_results}, the legacy fine-tuned Inception-ResNet system is limited, with 52.0\% Class Accuracy and 42.1\% Recall@1, contributing to weaker performance of the end-to-end Class-dependent system shown in \Cref{tab:main_results}.
In contrast, our fine-tuned OpenCLIP model achieves 79.6\% Top-1 Accuracy and 56.4\% Recall@1 in exact product retrieval, which is a 14.6 percentage point absolute improvement over the OpenCLIP baselines, demonstrating its ability to capture both broad category semantics and fine visual details for a robust, classification-free search pipeline.
\section{Case Study: Visual Search at Wayfair}

To demonstrate the practical utility and business impact of our taxonomy-decoupled architecture, we present a case study of its deployment on our global home goods e-commerce platform.
This section details the system's integration into multiple user-facing features and quantifies its real-world value by analyzing live business metrics following the full-scale production rollout.

\subsection{E-Commerce Platform Integrations}

Our visual search system is integrated into multiple touchpoints of the e-commerce platform to serve distinct user needs, from direct retrieval to open-ended discovery.
First, as a direct search tool, it allows users to submit an image query via the main search bar.
This enables a non-textual retrieval pathway for users who recognize an item visually, without requiring precise descriptive terms to convey their intent.
Second, as a contextual recommendation tool on product pages, the engine analyzes inspirational scene imagery and presents carousels of shoppable products that are aesthetically and functionally similar to those depicted.
This allows customers to deconstruct a complete look and find coordinated items.
Finally, our system powers an inspirational discovery tool that moves beyond item-level retrieval.
It leverages AI-generated scenes as visual queries to surface an infinite scroll of aesthetically related scenes.
This transforms the user experience from transactional item-finding to a serendipitous exploration of styles, fundamentally broadening the application of visual search.

\subsection{Live Customer Metrics}

To validate our proposed system's real-world impact, we analyzed key business metrics before and after its full-scale production deployment, which replaced the legacy Class-dependent system.
Our analysis reveals a multi-faceted positive impact, demonstrating gains in user engagement and the creation of new commercial opportunities.

Following deployment, we observed a substantial and sustained increase in product discovery.
The Product Detail Page (PDP) view rate, a key indicator of user engagement, improved from 46\% to 61.8\% in the direct search tool.
A similar lift was observed in the contextual recommendation tool on product pages, where the rate increased from 42\% to 46\%.
Interestingly, Add-to-Cart and Conversion Rates remained stable.
This indicates that the primary gains occurred in the upper funnel of the customer journey.
Our system excels at helping users explore and discover relevant products, even if they are not yet ready to make an immediate purchase, which directly aligns with the goal of capturing subjective, open-ended user intent.
This result empirically validates that the improvements in our offline metrics, particularly Visual Similarity Precision and Success Rate (\Cref{tab:main_results}), translate to a measurable increase in real-world user engagement.

Beyond improving existing functionalities, the system's architectural flexibility unlocked entirely new commercial opportunities.
To quantify the business impact of the new inspirational discovery feature enabled by the proposed system, we employed a synthetic control framework to compare user outcomes with and without exposure to the feature.
The analysis revealed a clear commercial uplift, with the expected Variable Contribution Dollars (eVCD) improving by 3.86\%.
This gain is particularly significant given that such a discovery mechanism was infeasible under the prior rigid, class-dependent architecture, highlighting a key advantage of our proposed class-agnostic system.

Overall, these results demonstrate that our system produces measurable gains in downstream business metrics, providing empirical validation of its effectiveness in aligning subjective user intent with both user engagement and commercial outcomes.
\section{Conclusions}

We presented a taxonomy-decoupled visual search architecture that separates object detection from fine-grained taxonomy classification, providing greater flexibility and adaptability.
We then proposed an LLM-as-a-Judge evaluation framework that addresses the brittleness and evaluation challenges of conventional systems.
Deployed to the production environment at Wayfair, we demonstrated that the offline metrics from our framework strongly correlate with significant uplifts in real-world user engagement and product discovery.
This validates our decoupled system design and establishes a reliable methodology to evaluate industrial visual search systems.
Looking ahead, we plan to extend this system into a multimodal discovery tool to further enhance users' discovery experience.
{
    \small
    \bibliographystyle{ieeenat_fullname}
    \bibliography{main}
}

\end{document}